# ReNikud: Audio-Supervised Hebrew Grapheme-to-Phoneme Conversion

https://renikud.github.io/

Maxim Melichov*
*Reichman University*

Yakov Kolani*
*Independent Researcher*

Morris Alper
*Carnegie Mellon University, University of Miami*

***Abstract*—Grapheme-to-phoneme (G2P) conversion for Modern Hebrew is needed for applications like text-to-speech (TTS), but is challenging due to the language's abjad writing system, which leaves vowels largely unwritten, creating substantial ambiguity. Standard approaches first predict vowel diacritics (nikud) to produce International Phonetic Alphabet (IPA) transcriptions, but this is limited: vocalization data is scarce and laborious to produce, it does not specify features such as lexical stress, and it reflects formal grammatical rules rather than everyday spoken pronunciation. Direct sequence-to-sequence IPA prediction, meanwhile, struggles on limited data and fails to exploit the character-level alignment characteristic of abjads. Our method, ReNikud, overcomes these limitations with two key insights: (1) Weak audio supervision via a phoneme-based automatic speech recognition (ASR) pseudo-labeling pipeline on thousands of hours of unlabeled Hebrew audio, yielding phonemic transcriptions that reflect natural spoken norms without manual annotation. (2) A pseudo-vocalization architecture that predicts IPA phonemes at each character position, enforcing character-level alignment as an inductive bias. Results on existing Hebrew G2P benchmarks and the new targeted MILIM benchmark for spoken Hebrew show that ReNikud surpasses previous state-of-the-art methods. We release our code and trained models to support further work on Hebrew TTS and speech technologies.**

## I. Introduction

With increasing interest in text-to-speech (TTS) systems for low-resource languages, the modern Hebrew language poses particular challenges. Hebrew is written as an *abjad*—a writing system that normally does not indicate vowel sounds [1]. Consequently, many words are homographs and require semantic context for correct pronunciation. For example, the word מספרים can be read as /mispaʁˈim/[1] (*numbers*), /mispaʁˈa-jim/ (*scissors*), or /mesapʁˈim/ (*telling*). As this ambiguity confounds TTS generations, leading open-source approaches first predict vowel diacritics (*nikud*) that clarify pronunciation to condition synthesis [2], [3]—for example, adding diacritics to the unvocalized consonants ספר distinguishes סֵפֶר /sefeʁ/ (*book*) from סַפָּר /sapaʁ/ (*barber*).

One issue is the mismatch between written and spoken Hebrew. Firstly, conventional diacritics do not fully specify phonetic features such as lexical stress, e.g., בירה may be pronounced as /b'iʁa/ (*beer*) or /biʁ'a/ (*capital city*). Secondly, these diacritics reflect traditional grammatical rules that do not match everyday spoken Hebrew. For instance, the word ובדרך is formally vocalized and pronounced as /uvadˈeʁeχ/, but in natural speech, it is pronounced /vebadˈeʁeχ/. Similarly, בירושלים is formally /biʁuʃalˈajim/, but native speakers typically say /bejeʁuʃalˈajim/ [4], [5].

Another issue is a lack of scalable data sources for directly learning G2P. Vocalization models [6], [7] learn from Hebrew text with manually annotated vowel diacritics, data which is scarce and laborious to produce. Similarly, methods such as [3], [8] that learn from IPA annotations are bottlenecked by data availability. Conversely, another abundant source of data on pronunciation exists—unlabeled Hebrew audio—but existing methods can only learn from text.

Finally, while methods using vocalization poorly reflect spoken pronunciation, they outperform direct sequence-to-sequence (seq2seq) prediction of IPA [3], the latter of which struggles to learn on limited data while failing to use character-level alignment with the input transcript as an inductive bias.

Our method, *ReNikud*[2], directly addresses these challenges. To reflect spoken norms and unlock abundant training data, we propose a novel pipeline to train G2P with *weak supervision from unlabeled Hebrew audio*. By applying an automatic speech recognition (ASR)-based pseudo-labeling pipeline, we are able to extract pronunciation information and train on thousands of hours of recordings, a scalable approach for learning spoken Hebrew norms. To enforce character-level alignment between Hebrew characters and phonemes, we introduce a *pseudo-vocalization* architecture. Like traditional vocalization, this predicts phonetic content aligned with each character position, but rather than predicting traditional diacritics it directly predicts character-aligned IPA phonemes. By using the orthographic structure of Hebrew text, this increases data efficiency relative to seq2seq baselines, as we demonstrate.

Our results on existing Hebrew G2P benchmarks show that ReNikud succeeds in leveraging audio data to better reflect actual Hebrew speech. Furthermore, because standard benchmarks predominantly evaluate formal text, they often fail to capture the complexities of modern spoken Hebrew.

*Equal contribution.

[1] We place the IPA stress mark directly before the stressed vowel, following TTS conventions.

[2] Rethinking Nikud

To address this critical gap, we introduce MILIM as a core contribution. This novel, targeted benchmark systematically evaluates difficult phonetic phenomena—such as slang, foreign loanwords, acronyms, and complex homographs—providing a much-needed framework for measuring how well models handle the nuances of contemporary spoken Hebrew.

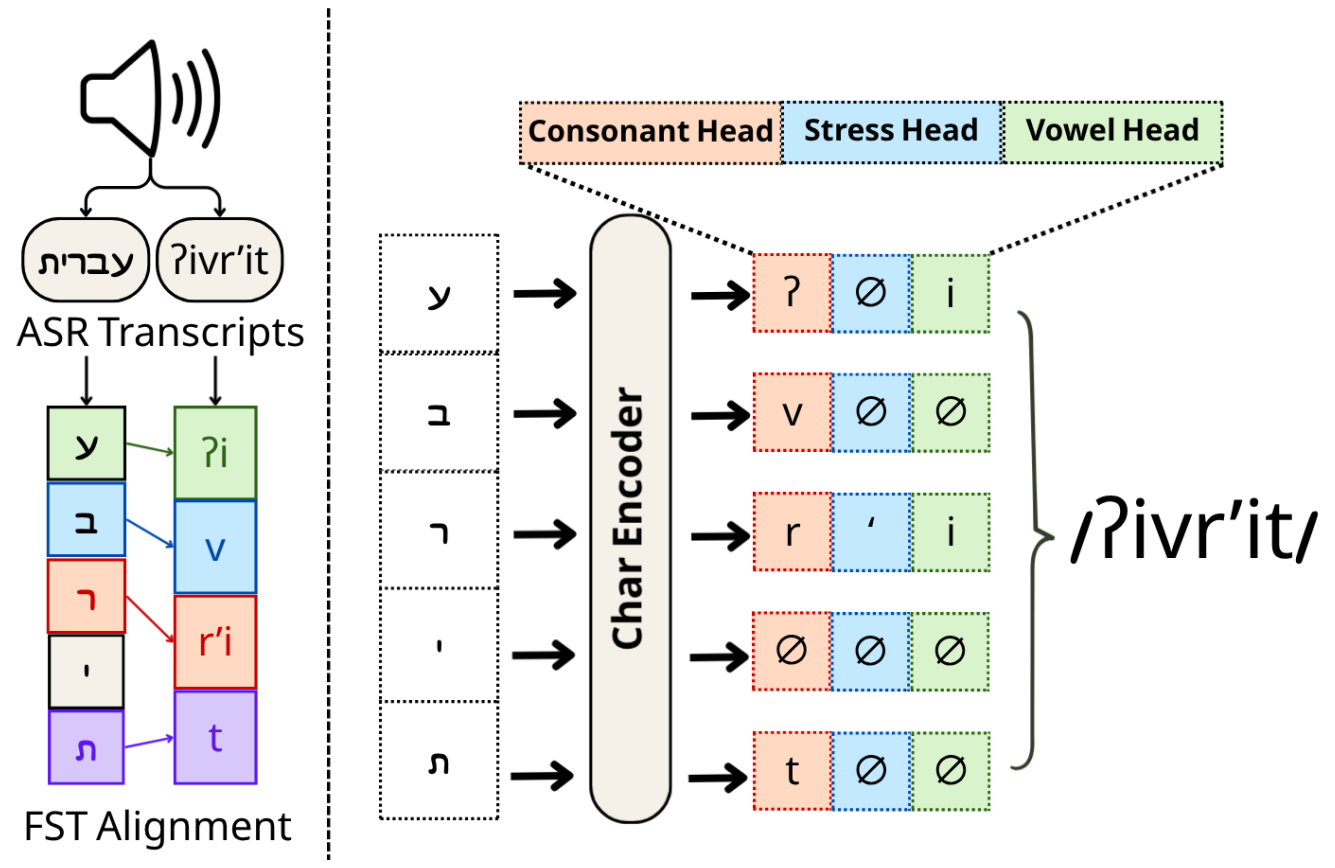

Figure 1: **System overview.** We first pseudo-label audio (left) by creating a many-to-one FST alignment between unvocalized Hebrew text and IPA phonemes derived from two parallel ASR runs applied to Hebrew audio. We then train a pseudo-vocalization architecture (right) where unvocalized Hebrew characters are passed through a character encoder to predict a phonetic triplet (consonant, stress, and vowel) at each position via parallel classification heads.

## II. Method

The ReNikud pipeline consists of two stages, shown in Figure 1: audio pseudo-labeling (Section II-A), and our pseudo-vocalization architecture trained on this data (Section II-B).

### *A. Audio Pseudo-Labeling*

To learn pronunciation from unlabeled audio at scale, we construct a pipeline that extracts character-aligned IPA annotations using two parallel ASR systems, as shown in Figure 1 (upper left). We extract Hebrew orthographic transcripts with a standard pretrained Hebrew ASR model, and IPA transcripts with a custom ASR model trained to output IPA when applied to Hebrew audio. By applying both ASR systems to a large-scale, unlabeled Hebrew audio corpus, we extract parallel Hebrew text and IPA transcriptions that serve as *pseudo-labels*, providing weak supervision for our downstream G2P model.

To find character-level correspondences between orthographic and IPA transcripts, we perform a string alignment process based on the Hebrew orthography's abjad structure. In general, abjads encode consonants linearly with interleaved, unwritten vowels, meaning that graphemes map monotonically to (consonant, vowel) pairs. As Hebrew also has unwritten lexical stress, each grapheme maps to a *phonetic triplet* encoding such a pair and a binary stress indicator. We find this alignment with a simple finite state transducer (FST) handling known consonant values, including one-to-many (e.g., ב→/b, v/) and many-to-one (e.g., ט, ת→/t/) mappings, as well as orthographic complexities such as:

- **Digraphs:** Loanwords in Modern Hebrew frequently use an apostrophe (*geresh*) to denote non-native phonemes (e.g., ’ז for /ʒ/; the base letter ז normally represents /z/). The FST assigns the digraph phoneme to the base letter and passes over the *geresh*.
- **Word-final ח** (/χ/) may occur with an additional *preceding* /a/ vowel (*patah gnuva*), e.g., לוח /l'uaχ/. We include a special combined /aχ/ value for the vowel slot.
- **Silent letters**: The letters ו, י, ה, א may either indicate consonant sounds or may be silent (*matres lectionis*). We handle the latter case with a null (∅) consonant class.

Examples of the resulting alignment are shown in Table I.

| Word | Char | Consonant | Vowel | Stress |
|---|---|---|---|---|
| שלום (/ʃal'om/) | ש | /ʃ/ | /a/ | 0 |
| | ל | /l/ | /o/ | 1 |
| | ו | ∅ | ∅ | 0 |
| | ם | /m/ | ∅ | 0 |
| צ׳׳פס (/tʃ'ips/) | צ | /tʃ/ | /i/ | 1 |
| | ׳ | ∅ | ∅ | 0 |
| | ׳ | ∅ | ∅ | 0 |
| | פ | /p/ | ∅ | 0 |
| | ס | /s/ | ∅ | 0 |
| תפוח (/tap'uaχ/) | ת | /t/ | /a/ | 0 |
| | פ | /p/ | /u/ | 1 |
| | ו | ∅ | ∅ | 0 |
| | ח | ∅ | /aχ/* | 0 |

Table I: **Examples of FST-derived alignment** between Hebrew characters and phonetic triplets (consonant, vowel, stress). *Special combined value for *patah gnuva*.

We filter for quality by checking for agreement between the orthographic and IPA transcripts, retaining only samples where both ASR systems produce matching word counts. In addition, we filter for successful FST alignment. Instead of discarding full samples with FST mismatches, we extract all maximal contiguous aligned word sequences. Consequently, a single unaligned word splits a larger sample into multiple valid, shorter samples. This fragmentation explains how the original 474K samples yielded 1.52M shorter samples, retaining approximately 60% of the total words.

Qualitative investigation of the ∼40% of discarded content suggests that it consists primarily of ASR transcription noise and speech artifacts. These include Hebrew ASR typos (e.g., transcribing חמשים instead of חבר against the IPA /χavˈeʁ/), IPA ASR truncations (e.g., aligning the word קודם with the truncated IPA /kˈod/), or misalignment due to speech disfluencies and stuttering.

### *B. Pseudo-Vocalization Architecture*

Our goal is to create a Hebrew G2P model that maps unvocalized Hebrew text directly to IPA strings. Because Hebrew is an abjad, there is a strong, local relationship between individual written letters and their phonetic realizations, which standard sequence-to-sequence models fail to exploit as an inductive bias.

To mitigate this, we frame the G2P task as a constrained, per-character classification problem—a method we term *Pseudo-Vocalization*, illustrated in Figure 1 (right). While a single Hebrew character typically corresponds to more than one IPA symbol (e.g., a consonant followed by a vowel), we resolve this by having every Hebrew letter independently predict exactly one phonetic triplet, as defined in Section II-A.

The core model is a character-level transformer encoder, with three parallel, independent classification heads that simultaneously predict the phonetic attributes for each character directly from the encoder's hidden states:

- **Consonant Head:** Selects from 25 IPA consonants or null ($\emptyset$).
- **Vowel Head:** Selects from 5 vowels, null ($\emptyset$), or the special /aχ/ token (see Section II-A).
- **Stress Head:** Binary classifier for lexical stress.

At inference time, realizations are predicted by taking the argmax of logits for each head. In addition, we apply *constrained decoding* to enforce hard constraints on Hebrew letters and phonetic realizations: the argmax is calculated only over possible consonantal realizations of a letter (e.g., ב can only be realized as /b/ or /v/). We also enforce a word-level constraint that exactly one lexical stress is predicted. Ablating these restrictions confirms their necessity: unconstrained decoding degrades performance, dropping to 24.9% WER and 7.1% CER (compared to our fully constrained results of 24.3% WER and 6.6% CER).

## III. Experiments and Results

### A. Experimental Setup

As our ASR model for outputting orthographic transcripts, we employ the Whisper Large v3 Turbo checkpoint fine-tuned on Hebrew by ivrit.ai [9]. For ASR prediction of IPA, we adapt this model with two fine-tuning stages: (A) We first train on IPA pseudo-labels produced by Phonikud [3] applied to transcripts from the *SASpeech* [10] (∼18h[3]) and *Recital* [11] (∼50h) audio corpora. (B) We then fine-tune on the train split of the *ILSpeech* [3] (∼2h) audio corpus, containing expert-annotated gold IPA transcripts. After training until convergence, this model achieves strong performance on held-out data (e.g., 2.4% Character Error Rate on the test set of ILSpeech) and in qualitative inspection of predictions, supporting its use in our pipeline.

Our G2P model is initialized with a `dicta-il/dictabert-large-char` encoder with three added classification heads as described in Section II-B. For training data, we apply our pseudo-labeling pipeline (Section II-A) to Knesset Vox [12], a corpus of ∼1.7K hours of parliamentary recordings. Although this data reflects a governmental setting, the recorded speech mostly reflects informal spoken Hebrew phonology[4], captured by our downstream pseudo-labels.

We re-segment the raw audio into 5–15 second clips, as ASR quality degrades on longer inputs, and process each clip through both orthographic and IPA ASR models to obtain parallel transcripts. After FST alignment and filtering for transcript agreement and length outliers ($1.5 \times$ IQR), the pipeline yields 1.52M aligned Hebrew-to-IPA sentences. Models are trained until convergence on a held-out validation set.

### B. MILIM Benchmark

We introduce the MILIM[5] benchmark, a corpus of Hebrew sentences with paired IPA annotations for targeted words in context. This is designed to assess Hebrew G2P models' ability to perform complex phonetic disambiguation in challenging Hebrew contexts, such as spoken norms differing from formal language. The benchmark contains eleven categories each consisting of 150–151 sentences containing one or more targeted words (totals: 1,653 sentences, 3,110 target words including ILSpeech). These were produced with a semi-manual procedure including manual production as well as verification and correction of items produced by Gemini-3.1-pro when shown existing items as in-context seeds. MILIM also includes the test split of ILSpeech as an additional control category (using version 2, with minor orthographic fixes from the original ILSpeech release).

Examples from MILIM are shown in Table II. The categories test the following challenges:

- **Gender:** Pronouns and suffixes whose pronunciation depends on the contextually licensed gender of the referent.
- **Homographs:** Identically spelled words with different pronunciations (excluding minimal stress pairs, below).
- **Slang:** Informal vocabulary.
- **Colloquial:** Words where everyday spoken pronunciation diverges from prescriptive norms.
- **Rare Phonemes:** Words with rare or non-native phonemes marked by geresh diacritics.
- **Foreign:** Loanwords and non-native vocabulary.
- **Acronyms:** Hebrew acronyms requiring correct stress and vowel resolution.
- **Penultimate Stress:** Words stressed on the penultimate syllable, unlike Hebrew's default final stress.
- **Minimal Stress Pairs:** Word pairs distinguished only by stress placement.
- **Names:** Proper nouns, which often have non-standard pronunciation patterns.
- **ILSpeech-v2:** The existing ILSpeech benchmark.

### C. G2P Evaluation

We evaluate all models on MILIM using word-level WER and CER against gold IPA annotations, macro- and micro-averaged across categories (Table III, Figure 2).

ReNikud outperforms the baseline, with the largest gains on categories reflecting spoken Hebrew norms. On `colloquial` items, ReNikud correctly predicts spoken forms such as ומשפחה $\rightarrow$ /vemiʃpaχˈa/, where both Phonikud and Gemini default to the prescriptive /umiʃpaxˈa/. On `slang`, ReNikud correctly resolves items like פאדיחה $\rightarrow$ /fadˈiχa/ (Phonikud: */padiχˈa/). On `foreign` items, ReNikud maps non-native phonemes correctly, e.g., ווינר $\rightarrow$ /wˈineʁ/ (Phonikud: */vinˈeʁ/). Additional examples are shown in Table V.

[3]Filtered from the original 26h to remove segments with filler words.

[4]Formal phonology is primarily used in scripted broadcasts and recitation of vocalized texts.

[5]Meaning "words" in Hebrew; this benchmark evaluates pronunciation at the word level.

| Cat. | Input | Target | IPA | Cat. | Input | Target | IPA |
|---|---|---|---|---|---|---|---|
| Gender | תגידי, הכל בסדר איתך? | איתך | /ʔitˈaχ/ | Names | ליהי איפה החולצה הכחולה שלי? | ליהי | /lˈihi/ |
| Acronyms | הוא שירת באמ"ן | באמ"ן | /beʔamˈan/ | Homogr. | הוא קרא ספר מעניין בזמן שהוא ישב אצל ספר נשים. | ספר | /sˈefeʁ/~/sapˈaʁ/ |
| Penult. | קניתי לחם טרי | לחם | /lˈeχem/ | Min. Str. | היה לו נחת רוח מההישגים של ילדיו / המטוס נחת בשלום על המסלול | נחת | /nˈaχat/~/naχˈat/ |
| Rare Ph. | הבאתי צ'יפס בשבילך, תאכל מהר. | צ'יפס | /tʃˈips/ | Slang | הוא בן אדם גנוב לגמרי. | גנוב | /ganˈuv/ |
| Foreign | איזה וויב טוב יש פה | וויב | /vˈajb/ | Colloq. | הכל תלוי בחברים ומשפחה תומכת. | ומשפחה | /vemiʃpaχˈa/ |
| ILSpeech | אני אוכל לעשות מאה ימי צילום בחודש. | *(Full Sentence)* | /ʔanˈi ʔuχˈal laʔasˈot mˈeʔa jemˈej tsilˈum beχˈodeʃ/ | | | | |

Table II: Selected examples from MILIM, split by category (Cat.). Abbreviated category names are penultimate stress (Penult.), rare phonemes (Rare Ph.), homographs (Homogr.), colloquial (Colloq.), and minimal stress pairs (Min. Str.).

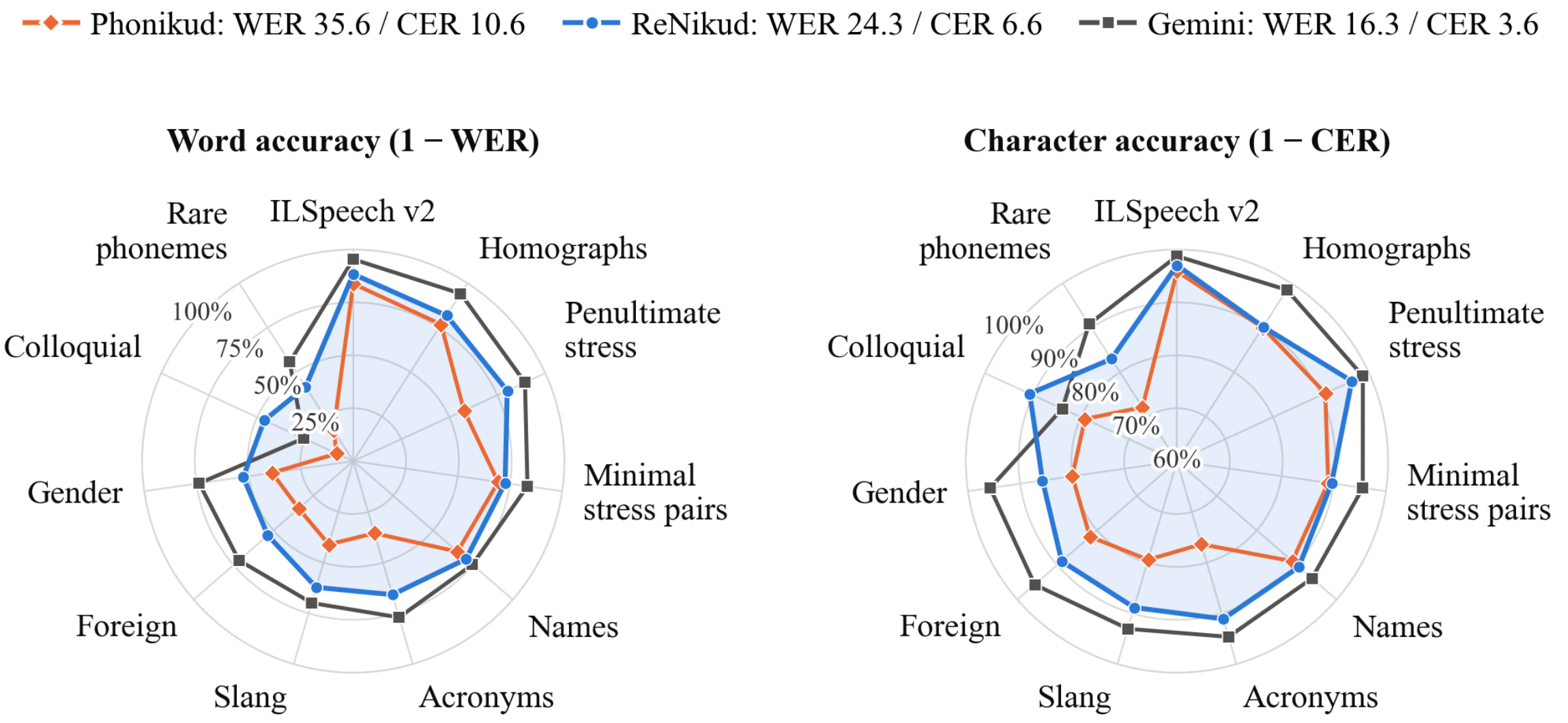

Figure 2: Word accuracy rate (1 – WER) (left) and character accuracy rate (1 – CER) (right). The black line represents Gemini 3.1 Pro, serving as a practical upper bound as the best-performing LLM.

We evaluate several frontier LLMs as reference upper bounds (Table III), evaluated on sentences batched into groups of 250 due to resource limitations, and with simple IPA normalization applied (e.g., collapsing /x/ and /χ/). Gemini generated the test sentences (excluding ILSpeech), so its scores may reflect upward bias; Claude-Fable-5, with no such role, achieves comparable performance. LLMs remain costly and impractical for real-time TTS, and our results show they struggle with colloquial forms attested only in audio, a limitation which would be inherited even with distillation.

### D. User Study of MILIM Benchmark

To verify the validity of MILIM for benchmarking spoken Hebrew pronunciation, we conduct a native speaker survey. We focus on the Colloquial category, which is poorly attested in written resources, exhibits the sharpest divergence between formal and spoken Hebrew and provides a natural binary comparison (spoken vs. formal pronunciation). We synthesize pronunciations for aural comparison via BlueTTS[6], a Hebrew IPA-to-speech engine. Nine native Hebrew speakers rated 150 audio pairs (informal vs. formal pronunciation, with written context) on a 7-point Likert scale indicating which sounded more like casual, spoken Hebrew. Each pair was evaluated by three raters; order was randomized and condition labels hidden. The results show a consistent preference for the informal variants, with a mean CMOS of $+0.87$ (std $= 1.50$) across 450 ratings from 9 native speakers (150 items). A linear mixed-effects model with crossed random intercepts for rater and item confirms this preference is statistically significant ($\beta = 0.87,\ p < 0.001$), confirming that annotations in MILIM match speaker intuitions of colloquial speech.

[6]https://github.com/maxmelichov/BlueTTS

| | Gender | Acronyms | Penult. Stress | Rare Phonemes | Foreign | Names | Stress Homog. | Slang | Colloquial | Min. Stress Pairs | ILSpeech-test | **OVERALL** | **MICRO** |
|---|---|---|---|---|---|---|---|---|---|---|---|---|---|
| Targets ($N$) | 152 | 152 | 151 | 151 | 155 | 150 | 203 | 156 | 151 | 150 | 1,539 | – | 3,110 |
| *Proprietary / Non-Realtime Models* | | | | | | | | | | | | | |
| Gemini | 26.3 / 4.2 | 23.0 / 5.4 | 10.6 / 1.3 | 44.4 / 9.2 | 28.4 / 4.3 | 25.3 / 6.2 | 5.9 / 1.5 | 30.1 / 7.0 | 74.2 / 16.1 | 16.7 / 4.4 | 4.6 / 1.2 | 26.3 / 5.5 | 16.3 / 3.6 |
| GPT-5.5 | 42.8 / 7.7 | 47.4 / 11.5 | 38.4 / 6.7 | 38.4 / 9.2 | 41.9 / 8.8 | 69.3 / 22.4 | 22.2 / 5.9 | 37.2 / 8.9 | 83.4 / 26.5 | 44.7 / 14.2 | 29.5 / 6.1 | 45.0 / 11.6 | 37.7 / 9.0 |
| Opus 4.8 | 19.7 / 5.4 | 30.9 / 6.7 | 19.9 / 2.9 | 39.7 / 8.0 | 42.6 / 11.5 | 20.0 / 6.9 | 12.8 / 3.9 | 31.4 / 7.8 | 85.4 / 24.3 | 20.7 / 6.0 | 39.0 / 7.2 | 32.9 / 8.2 | 35.3 / 7.9 |
| Claude-Fable-5 | 34.2 / 5.4 | 21.7 / 5.1 | 30.5 / 3.8 | 25.8 / 5.4 | 32.3 / 6.4 | 14.0 / 4.7 | 8.9 / 4.9 | 20.5 / 5.1 | 82.1 / 18.5 | 18.0 / 6.1 | 11.7 / 2.7 | 27.2 / 6.2 | 20.0 / 4.7 |
| *Dedicated G2P Models* | | | | | | | | | | | | | |
| ReNikud (Ours) | **47.4 / 14.3** | **34.2 / 8.9** | **19.9 / 3.6** | **58.3 / 17.0** | **46.5 / 11.1** | **29.3 / 9.5** | **18.2 / 9.8** | **37.8 / 11.2** | **53.6 / 9.3** | **27.3 / 10.4** | **11.6 / 3.0** | **34.9 / 9.8** | **24.3 / 6.6** |
| Phonikud | 61.2 / 20.0 | 64.5 / 23.6 | 42.4 / 9.1 | 82.8 / 27.9 | 65.8 / 18.2 | 34.7 / 11.1 | 23.2 / 10.0 | 59.0 / 20.6 | 91.4 / 20.8 | 30.7 / 11.1 | 16.2 / 4.1 | 52.0 / 16.0 | 35.6 / 10.6 |

Table III: Detailed G2P evaluation on MILIM (WER / CER, in %) alongside per-category target word counts. The last two columns give aggregate results: macro-average across categories (OVERALL), and micro-average over samples (MICRO), the latter heavily weighted towards the larger ILSpeech test set. **Bolding** denotes the best-performing dedicated G2P model.

### E. Transfer to Diacritization

Diacritization (*nikud* prediction) is closely related to G2P, as both require resolving per-character phonetic ambiguities. However, diacritic annotation requires specialist linguistic knowledge that most native speakers lack, making labeled data scarce and difficult to scale—unlike unlabeled audio, which is abundant. We hypothesize that phonetic representations learned from audio supervision can transfer to diacritization. We retain the pre-trained ReNikud encoder and replace the phonetic classification heads with a nikud prediction head, following the per-character methodology of DictaBERT [7].

We fine-tune on subsets of 1k, 2.5k, 5k, and 10k sentences from the Knesset corpus (processed via the Dicta Nakdan API), with 10k held out for validation. We evaluate on 100 manually corrected sentences from the Nakdimon test set [6] (originally consisting of automatic pseudo-labels with frequent inaccuracies), comparing against a DictaBERT baseline trained with identical hyperparameters and early stopping.

As shown in Figure 3, the ReNikud-initialized encoder converges significantly faster at extremely small data sizes (1k: 23.6% vs. 37.4% WER, $p < 0.001$), suggesting that audio-supervised pretraining provides useful representations for diacritization when labeled data is severely limited. The model maintains a statistically significant advantage up to 10k sentences (13.4% vs. 15.7% WER, $p < 0.001$).

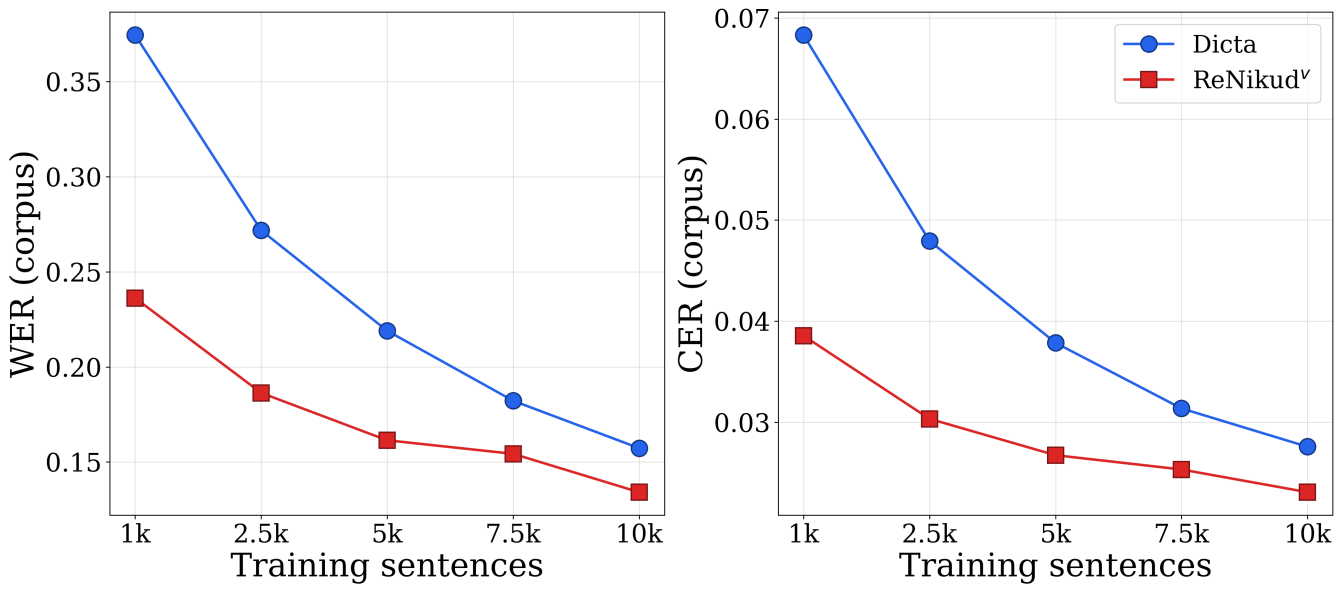

Figure 3: Diacritization performance on test set over the different number of sentences in the training.

### F. Ablations

We ablate both architecture and data source choices; per-category qualitative examples are in Table V.

**Architecture.** We compare our pseudo-vocalization heads against ByT5-Small (seq2seq) and a CTC network on the same DictaBERT encoder, all trained on identical Knesset Vox data (Table VI). Our method outperforms both overall. Without character-level alignment as an inductive bias, the seq2seq baseline frequently mispredicts vowels, stress placement, and other underspecified phonetic features—e.g., inserting a glottal stop in לאפה (/lˈafa/ → /leʔafˈa/) or defaulting to final stress on penultimate-stressed words (איפה /ʔˈejfo/ → /ʔejfˈo/). The CTC baseline shares our encoder but still underperforms, particularly on stress and vowel accuracy. To isolate the architecture from the seq2seq model choice, we also tested a seq2seq model using the DictaBERT encoder paired with a causal Transformer decoder and 3 linear heads, which suffered from the same alignment-related errors—for instance, incorrectly predicting epenthetic vowels (סמארטפון /smˈaʁtfon/ → /smˈaʁtafon/) and shifting stress on loanwords (פיצה /pˈitsa/ → /pitsˈa/).

**Data source.** We compare four data conditions with the same architecture. Comparing our full audio-supervised pipeline to training on two text-derived settings (Phonikud applied to Knesset Vox transcripts, and training on Phonikud's original text corpus), we observe that audio-derived labels outperform text-derived labels across all MILIM categories (Table III), with the largest gains on colloquial and slang items where spoken and written norms diverge most. We additionally train on ILSpeech's gold IPA alone (data used to fine-tune our IPA ASR teacher) and find a substantial degradation relative to our full pipeline, confirming the benefit of learning from large-scale audio.

| Model | צבא (Army) | כותרת (Title) | במומחה (An expert) | ומחקרי (And analytical) | לעתים (Sometimes) | פרסומות (Ads) | משכנתא (Mortgage) |
|---|---|---|---|---|---|---|---|
| ReNikud | צָבָא ✓ | כּוֹתֶרֶת ✓ | בְּמֻמְחֶה ✓ | וּמֶחְקָרִי ✓ | לְעִתִּים ✓ | פִּרְסוֹמוֹת ✓ | מַשְׁכַּנְתָּא ✓ |
| Dicta Baseline | צֶבָא ✗ | כֻּתֶּרֶת ✗ | בְּמֻוֹמְחָה ✗ | וּמִחְקָרִי ✗ | לְעָתִים ✗ | פִּרְסוּמוֹת ✗ | מַשְׁכַּנְתָא ✗ |

Table IV: Low-resource predictions (1k sentences). ReNikud's audio-supervised pretraining provides an inductive bias that guides correct vocalization, whereas the baseline model struggles with early-stage morphological and phonetic errors.

| Cat. | Word | Target | Ours | Phonikud | ByT5 |
|---|---|---|---|---|---|
| Gender | אצלך | /ʔetslˈeχ/ | ✓ | /ʔˈetslχ/ | /ʔetslˈχa/ |
| Homogr. | בצל×2 | /betsˈel/, /batsˈal/ | ✓, ✓ | /btsˈel/, /bˈatsal/ | /betsˈel/, /betsˈel/ |
| Colloq. | וברקים | /vebʁakˈim/ | ✓ | /uvʁakˈim/ | /ubʁakˈim/ |
| Slang | כפרה | /kapˈaʁa/ | ✓ | /kapaʁˈa/ | /kapeʁˈa/ |
| Acronym | במד"א | /bemˈada/ | ✓ | /bemadˈa/ | /bemadˈa/ |
| Penult. | קונספט | /kˈonsept/ | ✓ | /konsˈept/ | /konsˈept/ |
| Rare Ph. | הז'אנר | /haʒˈaneʁ/ | ✓ | /haʒanˈeʁ/ | ✓ |
| Foreign | האקר | /hˈakeʁ/ | ✓ | /hakˈeʁ/ | /haʔakˈeʁ/ |

Table V: Qualitative examples of G2P predictions on MILIM. ✓ = matches target. Common error patterns include stress misplacement (Phonikud, ByT5), formal conjunction defaults (Phonikud), and phoneme insertion (ByT5).

| Method | Val. (250) | | Test (3,110) | |
|---|---|---|---|---|
| | WER | CER | WER | CER |
| *Architecture comparison (all trained on Knesset Vox, audio-derived IPA):* | | | | |
| Seq2Seq (ByT5-Small)† | 24.1 | 4.9 | 32.1 | 11.0 |
| CTC (DictaBERT encoder) | 21.2 | 3.8 | 27.9 | 8.9 |
| Encoder-Decoder (DictaBERT) | 14.1 | 2.7 | 27.0 | 7.6 |
| Ours (Unconstrained) | 14.3 | 2.7 | 24.9 | 7.1 |
| **Ours (Constrained)** | **13.7** | **2.7** | **24.3** | **6.6** |
| *Data source comparison (all using our architecture):* | | | | |
| Gold ILSpeech IPA only | 33.0 | 7.1 | 40.8 | 12.8 |
| Phonikud text corpus | 23.4 | 6.8 | 32.6 | 11.0 |
| Knesset Vox, Phonikud-labeled IPA | 20.7 | 4.5 | 33.0 | 9.9 |
| **Knesset Vox, ASR-derived IPA (Ours)** | **13.7** | **2.7** | **24.3** | **6.6** |

†Uses ByT5 encoder; see text.

Table VI: Ablation G2P performance comparison. Error rates are reported as percentages for Word Error Rate (WER) and Character Error Rate (CER). The test set reflects the full MILIM suite. Best results in each comparison in **bold**.

## IV. Related Work

Explicit G2P conversion for Hebrew was recently introduced by Kolani et al. [3], following prior works on Hebrew diacritization [6], [7]. These approaches are bottlenecked by the availability of annotated textual data, while we use audio as a scalable source of pronunciation information.

Among prior works learning pronunciation from audio in other languages, we distinguish between two approaches:

(1) **Audio-supervised** methods such as ours use audio during training to improve pronunciation knowledge. Most similar to our work, a few studies use audio to improve G2P for English [13], [14] or in multilingual settings [15]. However, these works train on labeled audio as a supplementary signal to labelled text, while we use large-scale unlabeled audio as our primary training signal. Additionally, we operate on the Hebrew language which has a high degree of orthographic ambiguity and phonetic features unspecified in text (stress, spoken norms, etc.), and our pseudo-vocalization architecture uses abjad-style alignment as an inductive bias.

(2) **Audio-guided** methods also take audio along with text at inference time for G2P [16]–[18] or abjad diacritization [19], [20]. While adding audio as an additional input can provide a richer signal, we focus on the case where only text is available at inference time.

## V. Conclusion

We introduced ReNikud, a framework for Hebrew G2P conversion including both a *Pseudo-Vocalization* architecture and a methodology for leveraging audio data. We also propose the MILIM benchmark, on which evaluations show that ReNikud outperforms prior methods on Hebrew G2P—particularly in challenging cases such as colloquial, slang, rare phonemes, and penultimate stress. Finally, we show broader applicability to traditional diacritization, achieving competitive accuracy with substantially less training data than standard approaches.

**Limitations and Future Work:** A primary limitation of our methodology is its reliance on ASR-generated pseudo-labels, as inherent transcription errors from the ASR model naturally propagate into the G2P training data. Furthermore, because our FST alignment mechanism enforces a strict character-to-phoneme mapping, the aligner tends to discard highly informal spoken variants that deviate significantly from the written orthography (for example, when a speaker colloquially pronounces the first-person future tense verb אכתוב as /jixtov/).

Additionally, since the Knesset Vox dataset consists entirely of parliamentary speeches, the training corpus is skewed toward formal discourse; further work could incorporate more diverse sources of conversational audio. Finally, our framework shows promise for additional languages (i.e., Arabic) sharing challenges such as unwritten vowels and diverging written and spoken norms.

## Acknowledgements

Generative AI tools (GPT, Claude, Gemini) were used for polishing manuscript wording, formatting, and coding assistance.

## References

[1] P. T. Daniels and W. Bright, *The world's writing systems*. Oxford University Press, 1996.

[2] V. Pratap, A. Tjandra, B. Shi, P. Tomasello, A. Babu, S. Kundu, A. Elkahky, Z. Ni, A. Vyas, M. Fazel-Zarandi *et al.*, "Scaling speech technology to 1,000+ languages," *Journal of Machine Learning Research*, vol. 25, no. 97, pp. 1–52, 2024.

[3] Y. Kolani, M. Melichov, C. Calev, and M. Alper, "Phonikud: Hebrew grapheme-to-phoneme conversion for real-time text-to-speech," *arXiv preprint arXiv:2506.12311*, 2025.

[4] A. Aharoni, "Vocalization of modern hebrew," in *Encyclopedia of Hebrew Language and Linguistics*, G. Khan, S. Bolozky, S. E. Fassberg, G. A. Rendsburg, A. D. Rubin, O. Schwarzwald, and T. Zewi, Eds. Leiden: Brill, 2013, vol. 3, pp. 944–951.

[5] H. Neudecker, "Vocalization of modern hebrew and colloquial pronunciation," in *Encyclopedia of Hebrew Language and Linguistics*, G. Khan, S. Bolozky, S. E. Fassberg, G. A. Rendsburg, A. D. Rubin, O. Schwarzwald, and T. Zewi, Eds. Leiden: Brill, 2013, vol. 3, pp. 951–953.

[6] E. Gershuni and Y. Pinter, "Restoring hebrew diacritics without a dictionary," in *Findings of the Association for Computational Linguistics: NAACL 2022*, 2022, pp. 1010–1018.

[7] S. Shmidman, A. Shmidman, and M. Koppel, "Dictabert: A state-of-the-art bert suite for modern hebrew," 2023.

[8] J. Zhu, C. Zhang, and D. Jurgens, "Byt5 model for massively multilingual grapheme-to-phoneme conversion," in *Proc. Interspeech 2022*, 2022, pp. 446–450.

[9] ivrit.ai, "ivrit-ai/whisper-large-v3-turbo," https://huggingface.co/ivrit-ai/whisper-large-v3-turbo, 2025.

[10] O. Sharoni, R. Shenberg, and E. Cooper, "Saspeech: A hebrew single speaker dataset for text to speech and voice conversion," in *Proc. Interspeech*, 2023.

[11] ivrit.ai, "ivrit-ai/crowd-recital," https://huggingface.co/datasets/ivrit-ai/crowd-recital, 2025.

[12] Y. Marmor, A. Zulti, D. Krongauz, A. Gabet, Y. Snapir, Y. Lifshitz, and E. Segal, "Voxknesset: A large-scale longitudinal hebrew speech dataset for aging speaker modeling," 2026. [Online]. Available: https://arxiv.org/abs/2603.01270

[13] S. Sun, K. Richmond, and H. Tang, "Improving seq2seq tts frontends with transcribed speech audio," *IEEE/ACM Transactions on Audio, Speech, and Language Processing*, vol. 31, pp. 1940–1952, 2023.

[14] S. Sun and K. Richmond, "Acquiring pronunciation knowledge from transcribed speech audio via multi-task learning," 2024. [Online]. Available: https://arxiv.org/abs/2409.09891

[15] M. S. Ribeiro, G. Comini, and J. Lorenzo-Trueba, "Improving grapheme-to-phoneme conversion by learning pronunciations from speech recordings," *arXiv preprint arXiv:2307.16643*, 2023.

[16] J. Route, S. Hillis, I. C. Etinger, H. Zhang, and A. W. Black, "Multimodal, multilingual grapheme-to-phoneme conversion for low-resource languages," in *Proceedings of the 2nd Workshop on Deep Learning Approaches for Low-Resource NLP (DeepLo 2019)*, 2019, pp. 192–201.

[17] H. Gao, M. Hasegawa-Johnson, and C. D. Yoo, "G2pu: grapheme-to-phoneme transducer with speech units," in *ICASSP 2024-2024 IEEE International Conference on Acoustics, Speech and Signal Processing (ICASSP)*. IEEE, 2024, pp. 10 061–10 065.

[18] C.-J. Li, K. Chang, S. Bharadwaj, E. Yeo, K. Choi, J. Zhu, D. Mortensen, and S. Watanabe, "Powsm: A phonetic open whisper-style speech foundation model," *arXiv preprint arXiv:2510.24992*, 2025.

[19] S. Shatnawi, S. Alqahtani, and H. Aldarmaki, "Automatic restoration of diacritics for speech data sets," 2024. [Online]. Available: https://arxiv.org/abs/2311.10771

[20] A. Ghannam, N. Alharthi, F. Alasmary, K. Al Tabash, S. Sadah, and L. Ghouti, "Abjad ai at nadi 2025: Catt-whisper: Multimodal diacritic restoration using text and speech representations," pp. 757–761, 2025.